\documentclass[sigconf,natbib=true]{acmart}

\usepackage[ruled,linesnumbered]{algorithm2e}
\usepackage{multirow}
\usepackage{makecell}
\usepackage{enumitem}
\usepackage{caption}
\setitemize[1]{noitemsep,partopsep=0pt,parsep=0pt,topsep=0pt, leftmargin=10pt,
rightmargin=0pt}
\AtBeginDocument{%
  \providecommand\BibTeX{{%
    \normalfont B\kern-0.5em{\scshape i\kern-0.25em b}\kern-0.8em\TeX}}}

\setcopyright{acmcopyright}
\copyrightyear{2023}
\acmYear{2023}
\acmDOI{10.1145/1122445.1122456}

\acmConference[Woodstock '18]{Woodstock '18: ACM Symposium on Neural
  Gaze Detection}{June 03--05, 2018}{Woodstock, NY}
\acmBooktitle{Proceedings of the 32nd ACM International Conference on Information and Knowledge Management (CIKM),
  October 21--25, 2023, Woodstock, NY}
\acmPrice{15.00}
\acmISBN{978-1-4503-XXXX-X/18/06}

\begin{document}

\title{Counterfactual Monotonic Knowledge Tracing for Assessing Students' Dynamic Mastery of Knowledge Concepts}


\author{Moyu Zhang}
\affiliation{%
  \institution{Beijing University of Posts and Telecommunications}
  % \streetaddress{P.O. Box 1212}
  \city{Beijing}
  \state{Beijing}
  \country{China}
}
\email{zhangmoyu@bupt.cn}

\author{Xinning Zhu}
 \authornote{Corresponding Author}
\affiliation{%
  \institution{Beijing University of Posts and Telecommunications}
    \city{Beijing}
  \state{Beijing}
  \country{China}
}
\email{zhuxn@bupt.edu.cn}

\author{Chunhong Zhang}
\affiliation{%
  \institution{Beijing University of Posts and Telecommunications}
    \city{Beijing}
  \state{Beijing}
  \country{China}
}
\email{zhangch@bupt.edu.cn}

\author{Wenchen Qian}
\affiliation{%
  \institution{Beijing University of Posts and Telecommunications}
    \city{Beijing}
  \state{Beijing}
  \country{China}
}
\email{wenchen@bupt.edu.cn}

\author{Feng Pan}
\affiliation{%
  \institution{Beijing University of Posts and Telecommunications}
    \city{Beijing}
  \state{Beijing}
  \country{China}
}
\email{Pan_Feng@bupt.edu.cn}

\author{Hui Zhao}
\affiliation{%
  \institution{Beijing University of Posts and Telecommunications}
    \city{Beijing}
  \state{Beijing}
  \country{China}
}
\email{hzhao@bupt.edu.cn}

\begin{abstract}
As the core of the Knowledge Tracking (KT) task, assessing students' dynamic mastery of knowledge concepts is crucial for both offline teaching and online educational applications. Since students' mastery of knowledge concepts is often unlabeled, existing KT methods rely on the implicit paradigm of \emph{historical practice $\rightarrow$ mastery of knowledge concepts $\rightarrow$ students' responses to practices} to address the challenge of unlabeled concept mastery. However, purely predicting student responses without imposing specific constraints on hidden concept mastery values does not guarantee the accuracy of these intermediate values as concept mastery values. To address this issue, we propose a principled approach called Counterfactual Monotonic Knowledge Tracing (CMKT), which builds on the implicit paradigm described above by using a counterfactual assumption to constrain the evolution of students' mastery of knowledge concepts. Specifically, CMKT first assesses students' knowledge concept mastery value based on their historical practice sequences. Then, CMKT sets the answer of the most recent practice as the opposite of the actual answer and, based on this counterfactual answer, assesses the student's corresponding counterfactual knowledge mastery value. During the model training process, CMKT constrains the update of the student's knowledge states by ensuring that the two types of knowledge mastery values of students satisfy a fundamental educational theory, the monotonicity theory, to provide specific semantics for the assessed mastery values by the model. Moreover, we predict the student's responses to practices by directly measuring the difference between the student's knowledge concept mastery value and the difficulty of the target question, using labels of students' responses to assist model training. Finally, we conduct extensive experiments on five real-world datasets to demonstrate the superiority of CMKT over multiple baseline models. The code is available at https://github.com/zmy-9/CMKT.
\end{abstract}


\maketitle

\section{Introduction}
\begin{figure*}[t]
  \centering
  \includegraphics[width=\linewidth]{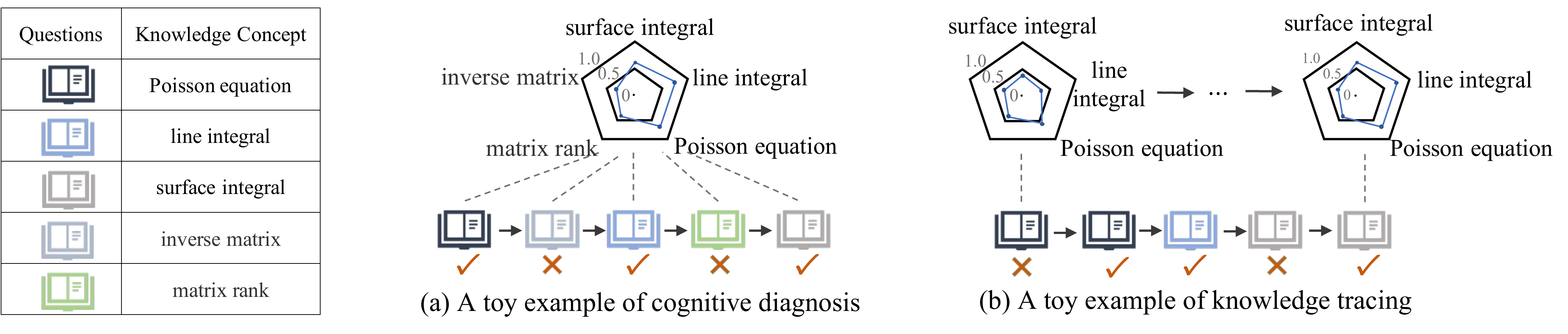}
  \caption{Examples of the cognitive diagnosis (CD) task and the knowledge tracing (KT) task. }
  \label{example}
  \vspace{-0.3cm}
\end{figure*}

Currently, there are two essential research tasks in online education: the Cognitive Diagnosis (CD) task \cite{cd_task1, cd_task2, irt} and the Knowledge Tracing (KT) task \cite{kt_task, bkt1, bkt2, kt_survey2, kt_survey3}. Although both tasks aim to track students' knowledge states based on their historical practice behavior, there are some distinctions between them. The CD task assumes that students' knowledge state is static and typically assesses their knowledge mastery in a testing environment (e.g., an exam), as depicted in Figure \ref{example}(a). In contrast, the KT task traces the dynamic changes in students' knowledge level and is generally employed for long-term student learning assessment, as shown in Figure \ref{example}(b). Assessing students' mastery of knowledge concepts in the KT task is more arduous than in the CD task because students' knowledge mastery evolves dynamically \cite{learning_efficiency1, learning_efficiency2}. Nevertheless, we endeavor to improve existing KT methods in assessing students' knowledge mastery, as evaluating students' evolving mastery of knowledge concepts can be of significant practical value in aiding online platforms to furnish understandable feedback and suitable learning materials to students \cite{kt_task, kt_survey, provide_materials}. For instance, by utilizing the assessed mastery values of knowledge concepts, we can generate personalized learning paths for students based on the educational knowledge graph on the online learning platform \cite{learn_path1, learn_path2, pre_relation1, pre_relation2}.

\begin{figure}[t]
  \centering
  \includegraphics[width=\linewidth]{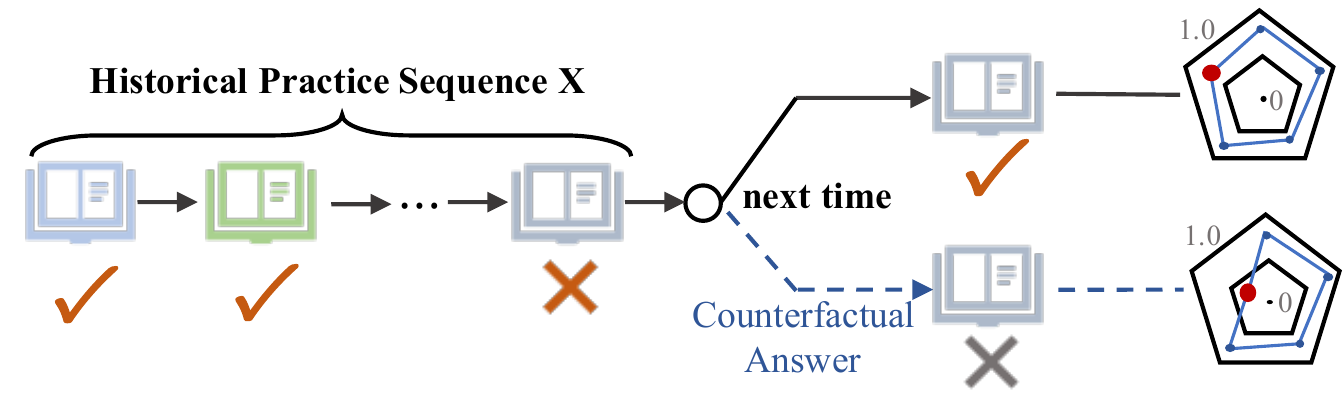}
  \caption{Illustration of the counterfactual assumption. }
  \label{mc_example}
\end{figure}

However, due to the lack of labels of students' mastery of knowledge concepts, both CD and KT tasks adopt the implicit paradigm of \emph{historical practices $\rightarrow$ mastery of knowledge concepts $\rightarrow$ students' responses to practices} to optimize the model's parameters, where the labels of students' responses are binary values indicating whether the student answered the question correctly. The initial motivation for this paradigm is reasonable, as students who have a higher master level of knowledge concepts are more likely to correctly answer questions than those who have a lower mastery level. However, in recent years, deep neural networks have been increasingly employed in KT models to enhance the accuracy in predicting students' responses \cite{dkt, akt, sakt, mf_dakt}, resulting in black-box properties of the models that make it challenging to ensure that the intermediate values in the model still reflect the students' mastery of concepts, thereby lacking interpretability and semantic information.

To more accurately track students' comprehension of concepts, Learning Process-consistent Knowledge Tracing (LPKT) \cite{lpkt} proposes to maintain consistency in the learning process by measuring learning gains based on the difference between current and past practices. However, LPKT only focuses on the positivity of the learning gain for each practice and fails to explicitly model the relationship between the relative size of learning gains for different practices. Consequently, LPKT cannot guarantee that the assessed mastery value of knowledge concepts satisfies the monotonicity theory \cite{monotonicity1}, which presumes that a student's mastery of a knowledge concept and the probability of incorrectly answering questions related to the concept is monotonous. There is a CD model called IRR \cite{irr}, proposes using a pairwise training method to directly incorporate this monotonicity during model training. However, the monotonicity design of IRR remains at the question granularity, rather than imposing monotonic constraints on the mastery of knowledge concepts. In fact, if we could directly constrain these mastery values to adhere to the monotonicity theory, we could give the assessed mastery values stronger interpretability and semantic information. However, applying the monotonicity theory directly to knowledge concept granularity is challenging, as we cannot be certain that a student who correctly answers a question involving multiple concepts has a higher mastery of each concept than a student who answers incorrectly, as the latter may only have a poor mastery level of a specific knowledge concept, not every concept. Consequently, imposing reasonable constraints on knowledge mastery values remains a significant challenge.

To address the above challenge, we propose a hypothesis: If a student's answer to the question answered in the previous practice are opposite to the true answer, what would be the difference between his knowledge mastery value and the mastery value of the real situation? Leveraging this counterfactual hypothesis, we propose a new model named Counterfactual Monotonic Knowledge Tracing (CMKT) to enhance the interpretability and rationality of assessing students' mastery of knowledge concepts. 

Specifically, as previously mentioned, the KT task requires modeling the update of students' knowledge state throughout the practice process, based on their historical practice sequences. Thus, CMKT devises a knowledge extraction module to model the student's practice sequence as their knowledge state, which is then mapped to the student's mastery value for each concept. Due to the lack of labels for students' knowledge mastery values, we employ the aforementioned implicit KT paradigm to train model parameter by predicting students' responses to practices. To ensure the knowledge mastery value's interpretability, we predict the student's response by directly comparing the relative size between the student's knowledge mastery value and the difficulty of the question based on the traditional CD method named Item Response Theory (IRT) \cite{irt, akt}. However, this method still optimizes the model at the question level, which does not guarantee that the mastery value of the specific concept satisfies the monotonicity theory. 

Considering that merely relying on predicting students' responses to practices does not ensure that the intermediate values of the model represent knowledge mastery values, we further propose a counterfactual monotonicity constraint during model training, as depicted in Figure \ref{mc_example}. Specifically, given a student's historical practice sequence, we can assess his mastery of concepts using KT models. However, if we assume that his most recent practice answer was the opposite of the true answer, assessed knowledge mastery value by the model in this counterfactual situation should differ from that assessed in the actual situation. For example, if the student's actual answer is correct, but we assume it is wrong, then based on the monotonicity theory, the mastery value of the relevant concepts in the counterfactual situation assessed by the model should not surpass the corresponding knowledge mastery value of the actual situation, and vice versa. Based on this counterfactual monotonicity assumption, CMKT constructs a counterfactual sample for each training sample during model training and generates sample pairs based on the counterfactual monotonicity for pairwise training, to constrain the concept mastery value in the model evaluation adhering to the monotonicity theory.

In addition, to evaluate the performance of KT methods in assessing students' dynamic mastery value of concepts directly and effectively, we propose a new metric called Group AUC of Mastery (GAUCM) based on the monotonicity theory. Finally, we employ the GAUCM metric along with two commonly used metrics for predicting students' responses to demonstrate the superior performance of CMKT compared to multiple baselines. In summary, the main contributions of this paper are as follows:
\begin{itemize}
\item To the best of our knowledge, we are the first KT study to propose counterfactual monotonicity. Our innovative approach combines counterfactual inferences with the monotonicity theory to explicitly constrain students' knowledge mastery at the knowledge concept level, making the mastery value update process more semantically informative and interpretable.
\item We define a structure to map students' historical practice sequences to their mastery values for each knowledge concept, and explicitly model the relationship between students' mastery values for knowledge concepts and their practice responses.
\item We conduct extensive experiments on five real-world KT datasets to validate the superiority of CMKT over state-of-the-art methods in assessing the value of students' mastery of knowledge concepts based on multiple metrics.
\end{itemize} 
 
\section{Related Work}
Current knowledge tracing (KT) methods can be divided into traditional methods and deep learning-based methods, where traditional types include Bayesian methods and factors-based methods. The most representative Bayesian method is Bayesian Knowledge Tracing (BKT) \cite{bkt1, bkt2}, which describes the knowledge state with a binary variable based on the Hidden Markova Model (HMM). The other line is the factor analysis method, where Item Response Theory (IRT) \cite{irt} is one of the most classical methods. IRT aims to model the relationship between students' ability and responses by measuring the gap between their ability and question difficulty, which has been applied in lots of CD models \cite{dirt, neuralcdm, cdmfkc, kscd},  and KT models. Later, Multidimensional IRT (MIRT) \cite{mirt} extends the parameters in IRT into multidimensional space. However, these methods ignore that the students' learning process is dynamic. Later, many models extended IRT by introducing students' practical behavior and many auxiliary information, such as AFM \cite{afm} and PFA \cite{pfa} record the number of students' practice to introduce practice information, DAS3H \cite{das3h} and Knowledge Tracing Machine (KTM) \cite{ktm} model introduce Factorization Machine (FM) \cite{fm} to construct a generalization framework that introduces richer features. MF-DAKT \cite{mf_dakt} introduces the deep neural network to further enhance the modeling capabilities of traditional methods. However, while these methods improve the predictive performance of the model, the interpretability of the model is also gradually weakened. 

In recent years, with the development of deep learning, most KT research mainly focuses on deep methods \cite{sakt, akt}. Deep Knowledge Tracing (DKT) \cite{dkt} firstly applies the LSTM \cite{lstm} into the KT task. Later,  the Memory-Augmented Neural Network (MANN) \cite{mann}, the Transformer structure \cite{transformer}, the Convolutional Neural Network (CNN)  \cite{cnn}, the Hawkes process \cite{hawkes} and other advanced network structures were introduced into the KT task to enhance the ability to predict students' responses to practices \cite{dkvmn,  sakt, akt, hawkeskt, ckt, iekt, dimkt, lfbkt}. To obtain students' concept mastery values, Dynamic Key-Value Memory model (DKVMN) \cite{dkvmn} utilizes a matrix to store students' dynamic mastery of concepts based on the MANN. Later, Graph-based KT method (GKT) \cite{gkt} and Structure-based KT method (SKT) \cite{skt} enhances DKT by exploiting the graph structure of concepts and directly models the students' mastery of each concept. Convolutional Knowledge Tracing method (CKT) \cite{ckt} applies the CNN to output a matrix representing students' mastery of concepts. Learning Process-consistent method (LPKT) \cite{lpkt} further proposes to focus on the evolution of students' states during the learning process by maintaining consistency of students’ changing knowledge state. However, although these methods also try to obtain students' concept mastery values, they generally represent the students' mastery of a concept in the form of vectors, and do not generate a mastery value during model optimization, making it difficult to ensure that these values conform to the monotonicity theory.  Therefore, in this paper, we propose the Counterfactual Monotonic Knowledge Tracing model to address the above issues.

\section{Preliminary}
In this section, we formally define the task of assessing students' mastery of concepts and briefly introduce the terminologies and the counterfactual monotonicity constraints proposed in this paper.
\subsection{Problem Statement}
For student $s$ at time $t$, we can observe his time-ordered answer history behavior before time t, denoted as $ X = \left\{(q_1, a_1), ..., (q_{t}, a_{t}) \right\} $, where $a_{t} \in \left\{ 0, 1 \right\} $ indicates whether $s$ answered question $q_t$ correctly. As a question may involve multiple knowledge concepts, we denote the concept set of $q_t$ as $K_{q_t}$. The embedding vectors for question $q_t$ and concept $k_j$ are denoted as $ \boldsymbol{q}_i\in \mathbb{R}^{d_q}$ and $\boldsymbol{k}_j \in \mathbb{R}^{d_k}$, respectively, where $d_q$ and $d_k$ denote the dimensionality of vectors.

Previous studies have highlighted the impact of question difficulty on student learning since correctly answering more challenging questions may significantly impact their knowledge states \cite{dimkt, mf_dakt}. Thus, we incorporate difficulty information into our model to model the learning process of students. Typically, the lower the correct answer rate, the more difficult the question is perceived by students. Based on this, we define the difficulty of a question as the ratio of incorrect answers to that question in the dataset, denoted as $\theta_{q} \in [0, 1]$. To create a sparse difficulty level feature, we multiply $\theta_{q}$ by a constant $C_q$ and round it, and then embed it as $\boldsymbol{\theta}_{q} \in \mathbb{R}^{d_{q{\theta}}}$ in a $d_{q{\theta}}$-dimensional vector. Similarly, we compute the average difficulty level for each knowledge concept across all questions, denoted as $\theta_{k} \in [0, 1]$ and use the constant $C_k$ to embed $\theta_{k}$ as $\boldsymbol{\theta}_{k} \in \mathbb{R}^{d_{k{\theta}}}$, where $d_{k{\theta}}$ is the dimension of the vector.

In this way, the practice record occurring at time $t$ can be formed as $p_t = (a_t, q_t, \theta_{q_t}, \left\{ k_i \right\}_{k_i \in K_{q_t}}, \left\{ \theta_{k_i} \right\}_{k_i \in K_{q_t}})$. We characterize this practice behavior by combining the raw embedding vectors of the input features. The specific calculation formula is as follows:
\begin{equation}
\boldsymbol{p}_t = \boldsymbol{W}_1 ^ T[\boldsymbol{a_t} \oplus \boldsymbol{q_t} \oplus \boldsymbol{\theta}_{q_t} \oplus \sum_{k_i \in K_{q_t}}(\boldsymbol{W}_0 ^ T(\boldsymbol{k}_i \oplus \boldsymbol{\theta}_{k_i}) + \boldsymbol{b}_0)] +  \boldsymbol{b}_1 ^ T
\end{equation}
where $\boldsymbol{a_t} \in  \mathbb{R}^{d_a}$ is the embedding of $a_t$. $\oplus$ is the concatenation operation. $\boldsymbol{W}_0 ^ T(\boldsymbol{k}_i \oplus \boldsymbol{\theta}_{k_i}) + \boldsymbol{b}_0$ denotes the combination of embedding of each concept in $K_{q_t}$ and corresponding concept difficulty. $\boldsymbol{W}_0 \in \mathbb{R}^{(d_k + d_{k{\theta}}) \times d}$, $\boldsymbol{W}_1 \in \mathbb{R}^{(d_a + d_q + d_{q{\theta}} + d) \times d}$ are the weight matrices, $\boldsymbol{b}_0 $, $\boldsymbol{b}_1 \in \mathbb{R}^d$ are the bias terms. $d$ is the dimension of vectors. 

\textbf{\emph{Task Definition}}. Given a student's historical practice sequence $X$ before time $t$, we aim to assess the student's mastery value of knowledge concepts after this period of practice, denoted as $ \boldsymbol{M}_t =\left\{m_{t1},  ...,  m_{ti}, ..., m_{t N_k}\right\}$. $N_k$ denotes the total number of knowledge concepts in the dataset. Here, $m_{ti}$ represents the student's mastery of knowledge concept $k_i$ at time t. As we cannot obtain labeled data on a student's mastery of a concept, we introduce the label of the student's responses to practices as a training objective and optimize the model parameters by predicting the student's future performance at the next time step $t+1$. Meanwhile, in this process, we hope that the student's knowledge mastery values of concepts, which are implicit intermediate values of the model, retain the property of knowledge mastery value.
\subsection{Terminology Definition}

To facilitate understanding of the process of assessing mastery of concepts in this paper, we first define the following terminologies:

\emph{Definition 3.1. (Students' Knowledge States)}. As students practice over time, their knowledge state changes as they receive learning gains with each practice. The knowledge states not only reflect their mastery of knowledge concepts at each moment, but also affect their subsequent performance. By modeling a student's historical practice sequence before time $t$ chronologically, we can obtain his knowledge state denoted as $\boldsymbol{h}_t = \boldsymbol{F}_{ke}(p_1, p_2, ..., p_t)$, where $\boldsymbol{h}_t \in \mathbb{R}^{d}$ and $d$ is the dimension of this hidden vector. $\boldsymbol{F}_{ke}$ represents a function that models knowledge state updates in a student's practice sequence, e.g., $\boldsymbol{F}_{ke}$ is the LSTM in the DKT model \cite{dkt}.

\emph{Definition 3.2. (Students' Mastery of Concepts)}.
Typically, KT studies set a student's mastery of a target concept on a scale of 0 to 1, where values closer to 1 denote greater levels of knowledge mastery \cite{dkvmn, dkt, lpkt, dkt, ckt, akt}. When a student attains a mastery value of 1 for a concept, he can be deemed to have fully mastered that concept. Assuming that a question pertains to only one knowledge concept, it can be inferred that students should be able to answer any related question correctly. And if a student can correctly answer the most challenging questions related to a particular concept, he can correctly answer all other related questions. Therefore, a student's mastery of a concept can be considered as the probability of accurately answering the most difficult practices related to that concept. To represent practices that involve the highest difficulty level of the target concept, we combine the embedding of the concept with the embedding of the highest difficulty level. We then calculate a student's mastery of concept $k_i$ at time $t$ as $m_{ti} = P(a=1|\boldsymbol{h}_t, k_{i}, C_{k})$,  where $C_{k}$ denotes the highest difficulty level of concepts.

\subsection{Counterfactual Monotonicity}
As previously mentioned, KT methods assess students' mastery of concepts by predicting their responses to practices, but these mastery values lack specific semantics due to the lack of direct constraints. We believe that the primary issue is the failure to make the mastery value conform to some fundamental educational theories, such as the monotonicity theory \cite{monotonicity1}. The theory asserts that students who answer a question correctly have a higher degree of mastery of concepts related to that question than students who answer incorrectly. Based on this, IRR \cite{irr} employed pairwise training loss to introduce the monotonicity into model training. However, the above monotonicity theory only works at the granularity of the question. For questions involving multiple knowledge concepts, students who answer the question correctly may not necessarily have a higher degree of mastery of each knowledge concept related to the question than students who answer incorrectly. A student may not possess a high level of mastery of a particular knowledge concept, resulting in an incorrect answer, but still display a high degree of mastery of other knowledge concepts. Therefore, faced with this complex student learning situation, achieving monotonic constraints at the granularity of concepts poses a challenge for us.

Since there are no labels for students' mastery of knowledge concept, it is challenging to verify whether concept-level mastery values vary monotonically based on existing samples. However, inspired by counterfactual learning \cite{counterfactual1, counterfactual2}, we propose that a monotonicity theory based on the counterfactual assumption. For instance, if a student answers a question $q$ related to two concepts $k_1$ and $k_2$ correctly, we denote that his mastery value for $k_1$ and $k_2$ after answering $q$ are $m_1$ and $m_2$, respectively. We then pose a counterfactual question: If he answers $q$ incorrectly, how would his mastery values for the two knowledge concepts change? Although we cannot determine the exact learning gain for answering $q$ correctly, we know that the state corresponding to a correct answer should have a higher mastery value than the value after an incorrect answer. Therefore, we assume that the mastery values after incorrect answers are $\overline{m}_1$ and $\overline{m}_2$, respectively. Instead of using uncertainty counterfactual inference, we can establish a simple and explicit counterfactual monotonic relationship: $m_1 \geq \overline{m}_1$ and $m_2 \geq \overline{m}_2$. Based on this counterfactual monotonic relationship, we can develop training objectives to help the model directly constrain the evolution of knowledge mastery during training, thereby obtaining more accurate and semantically meaningful mastery values.

\section{Method}
\begin{figure*}[h]
  \centering
  \includegraphics[width=\linewidth]{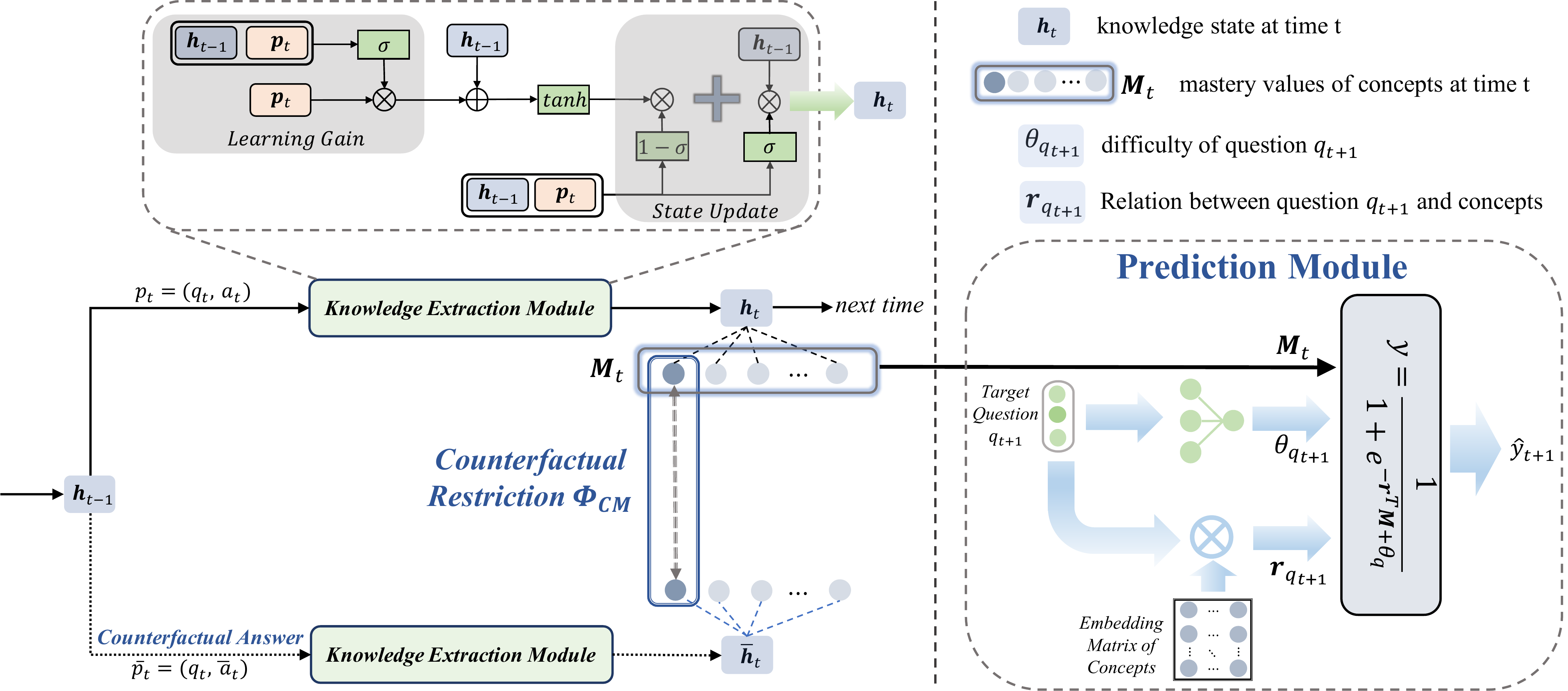}
  \caption{The architecture of the Counterfactual Monotonic Knowledge Tracing model (CMKT).}
  \label{model}
  \vspace{-0.3cm}
\end{figure*}

In this section, we provide details about the CMKT model. The primary structure of CMKT is shown in Figure \ref{model}. Specifically, CMKT employs a knowledge extraction module to model the evolution of students' knowledge states during practice. Then, based on \emph{Definition 3.2}, CMKT maps the hidden vectors that represent the knowledge states to their mastery values for each knowledge concept. We introduce a counterfactual monotonicity constrained regularization term to restrict the update process of students' mastery values of knowledge concepts, providing rich semantic information and enhancing the accuracy of the assessed mastery values by ensuring that knowledge mastery values satisfy counterfactual monotonicity. Moreover, to improve interpretability, we predict students' answers to questions based on the relative size between their knowledge mastery and the difficulty of the question, and then jointly optimize model parameters with a loss function for predicting student responses and the counterfactual monotonicity constraint term.

\subsection{Knowledge Extraction}
As a student's knowledge state changes with practice, we must update their knowledge states according to the sequence of their practice behaviors. Since each practice covers different questions with varying difficulty levels, the learning gains also differ. Therefore, we model the learning gains brought by each practice and update the knowledge states accordingly. The process is as follows: 

\emph{Learning Gain.} Students possess varying prior knowledge states, which can be seen as their learning abilities \cite{ckt}. Consequently, even when practicing the same question, students may experience different learning gains based on different prior knowledge states. Similar to previous works DIMKT \cite{dimkt} and LFBKT \cite{lfbkt},  we propose to model the learning gain of the practice $p_t$ using both the prior knowledge state and current practice-related features as:
\begin{gather}
\boldsymbol{z}_{t} =  \sigma(\boldsymbol{W}_2 ^ T(\boldsymbol{h}_{t-1} \oplus \boldsymbol{p}_t)+\boldsymbol{b}_2) \\ 
\boldsymbol{l}_{t} =  \boldsymbol{z}_{t} \odot \boldsymbol{p}_t \\
\boldsymbol{\widetilde{h}}_t = tanh(\boldsymbol{W}_3^T( \boldsymbol{h}_{t-1} \oplus \boldsymbol{l}_t) + \boldsymbol{b}_3)
\end{gather}
where $\sigma$ denotes the sigmoid function. $\odot$ denotes the element-wise product. $\boldsymbol{z}_{t} \in \mathbb{R}^{d}$ is the student's personalized perception of this practice under the prior knowledge state $\boldsymbol{h}_{t-1}$.  $\boldsymbol{l}_{t} \in \mathbb{R}^{d}$ represents the unique learning gain brought by $p_t$ under the prior knowledge state. Learning gain representations. We then can obtain the student's updated knowledge acquisition as $\boldsymbol{\widetilde{h}}_t \in \mathbb{R}^{d}$. $\boldsymbol{W}_2$, $\boldsymbol{W}_3 \in \mathbb{R}^{2d \times d}$ are the weight matrices,  and $\boldsymbol{b}_2$, $\boldsymbol{b}_3 \in \mathbb{R}^{d}$ are the bias terms. 

\emph{State Updating.} As students learn from practices, they may forget knowledge from the previous knowledge state due to the forgetting behavior that is closely related to both their current exercises and prior knowledge state \cite{dkt, das3h}. Thus, we need to model the forgetting rate at this moment in order to update their knowledge state:
\begin{gather}
\boldsymbol{f}_t = \sigma(\boldsymbol{W}_4^T(\boldsymbol{h}_{t-1} \oplus \boldsymbol{p}_t) + \boldsymbol{b}_4)  \\
\boldsymbol{h}_t = \boldsymbol{f}_t \odot \boldsymbol{h}_{t-1} + (\mathbf{1}- \boldsymbol{f}_t) \odot \boldsymbol{\widetilde{h}}_t
\end{gather}
where $\boldsymbol{h}_t$ denotes the updated knowledge state at time step $t$. $\boldsymbol{W}_4 \in \mathbb{R}^{2d \times d}$ is the weight matrix, and $\boldsymbol{b}_4 \in \mathbb{R}^{d}$ is the bias term. 

\subsection{\textbf{Assess Students' Mastery of Concepts}}
After obtaining a student's knowledge state, CMKT needs to map this hidden vector $\boldsymbol{h}_t$ to the mastery values of each knowledge concept. Based on \emph{Definition 3.2}, a student's mastery of a target concept can be defined as the probability of correctly answering the question that involves that concept with the highest difficult level. Therefore, we propose to combine the target concept embedding with the embedding corresponding to the highest difficulty level of concepts to represent a virtual question, which can then be used to measure a student's mastery of the concept, as follows: 
\begin{equation}
\boldsymbol{\widetilde{q}}_{k_i} =  tanh(\boldsymbol{W}_{5}^T (\boldsymbol{k}_{i} \oplus \boldsymbol{\theta}_{C_{{\theta}k}}) + \boldsymbol{b}_{5}) 
\end{equation}
where $\widetilde{q}_{k_i}$ denotes the virtual question which involves $k
_i$ with the highest difficult level. $\boldsymbol{W}_{5} \in \mathbb{R}^{(d_k + d_{{\theta}k}) \times d}$ is the weight matrix, and $\boldsymbol{b}_{5} \in \mathbb{R}^{d}$ is the bias term. At this time, due to that $\boldsymbol{\widetilde{q}}_{k_i}$ contains the highest difficulty level of $k_i$, a student can be seen as a higher probability of fully master concept $k_i$ when he has a higher probability correctly answering $\boldsymbol{\widetilde{q}}_{k_i}$. Then, we can calculate the students' mastery of concept $k_i$ based on $\boldsymbol{\widetilde{q}}_{k_i}$, i.e., $m_{ti}$, as follows: 
\begin{equation}
m_{ti} =  \sigma(\boldsymbol{W}_{5}^T (\boldsymbol{\widetilde{q}}_{k_i} \oplus \boldsymbol{h}_t))
\end{equation}
$\boldsymbol{W}_{5} \in \mathbb{R}^{2d \times 1}$ is the weight matrix used to predict the probability of correctly answering the virtual question based on the current knowledge state. In this way, we can obtain the student's mastery values of $K$ concepts at time $t$ as $\boldsymbol{M}_t =\left\{m_{t1},  ...,  m_{ti}, ..., m_{tK}\right\}$.

\subsection{\textbf{Counterfactual Monotonicity Restriction}}
Although there are some methods that have attempted to introduce monotonicity theory or impose constraints to regulate students' knowledge mastery mastery \cite{irr, lpkt}, none of them can enforce monotonic constraints at the knowledge concept granularity level for students' mastery of concepts. To bridge this gap, we use the counterfactual monotonicity theory proposed in Section 3.3 to directly constrain the mastery of each knowledge concept. 
Specifically, suppose a student's knowledge mastery level at a certain point is $M$, and his most recent practice was to answer question $q$ with the answer $A$. In the KT task, we can observe only one of two potential knowledge mastery outcomes for each student practice, depending on the student's response to question $q$ during that practice. If $A=a$, where $a$ denotes the real answer, we can observe a mastery level of $M=\boldsymbol{M}_{A=a}$. If $A=1-a$, we can infer a counterfactual result of $M=\boldsymbol{M}_{A=1-a}$. At this point, we need to learn the following functions:
\begin{gather}
\begin{aligned}
&
 \mathbb{E}[\boldsymbol{M}_{A=a} - \boldsymbol{M}_{A=1-a}|\boldsymbol{h}, q, A] \\
  = &\mathbb{E}[\boldsymbol{M}_{A=a}|\boldsymbol{h}, q, A] - \mathbb{E}[\boldsymbol{M}_{A=1-a}|\boldsymbol{h}, q, A]
  \end{aligned}
 \end{gather}
where $\boldsymbol{h}$ denote the student's knowledge state before answering $q$. According to the counterfactual monotonicity, we can deduce that if $a=1$, then $\mathbb{E}[\boldsymbol{M}_{A=a} -\boldsymbol{M}_{A=1-a}|A, q] \geq 0$,  otherwise $\leq 0$. The above formula can be interpreted as follows: Given a student's previous knowledge state $\boldsymbol{h}$ indicating the ability to correctly answer the next question $q$, and assuming the true answer $a$ is also 1, the student's knowledge level may remain the same or slightly improve. However, if we suppose that the student answered $q$ incorrectly, it can be inferred that the student has forgotten some knowledge, resulting in a decline in his mastery of concepts related to $q$. Thus, based on this counterfactual answer, the student's knowledge mastery value of concepts related to $q$ should be lower than the mastery value from the true answer. By applying the counterfactual constraint, we can rationalize the process of updating students' understanding of concepts. Specifically, for the practice $p_t$ that occurred at the previous time $t$, we can construct counterfactual samples as follows:
\begin{equation}
\boldsymbol{\overline{p}}_{t} = \boldsymbol{W}_1 ^ T[\boldsymbol{q_{t}} \oplus \theta_{q_{t}} \oplus \sum_{k_i \in K_{q_{t}}}(\boldsymbol{W}_0 ^ T(\boldsymbol{k}_i \oplus \boldsymbol{\theta}_{k_i}) + \boldsymbol{b}_0) \oplus \boldsymbol{\overline{a}}_{t}] +  \boldsymbol{b}_1 ^ T
\end{equation}
where $\boldsymbol{\overline{a}}_{t}$ denotes the embedding of the counterfactual answer $\overline{a}_{t}$, and $\overline{a}_{t} = 1- a_{t}$.  After that, we can obtain the student's counterfactual mastery of concept $k_i$, i.e., $\overline{m}_{ti}$, as follows:
\begin{gather}
\boldsymbol{\overline{h}}_{t} = \boldsymbol{F}_{ke}(\boldsymbol{h_{t-1}}, \boldsymbol{\overline{p}}_{t}) \\ 
\overline{m}_{ti} = m_{ti, \, A=1-a_{t}} =  \sigma(\boldsymbol{W}_{5}^T (\boldsymbol{\widetilde{q}}_{k_i} \oplus \boldsymbol{\overline{h}}_{t}))
\end{gather}
where $\boldsymbol{F}_{ke}(\cdot)$ denotes the knowledge extraction module. We then set restrictions to regularize the evolution of mastery of concepts \cite{hinge1, hinge2}, as shown by the following formula:
\begin{gather}
\begin{aligned}
\Phi_{CM} &=
\begin{cases}
 \mathbb{E}[\boldsymbol{M}_{A=a} -\boldsymbol{M}_{A=1-a} \, | \, A \, ] \geq 0, \quad a = 1 \\
 \mathbb{E}[\boldsymbol{M}_{A=a} - \boldsymbol{M}_{A=1-a}\, |\, A\, ] \leq 0, \quad a = 0
\end{cases}
 \\
&=\sum_{t=1}^T \sum_{k_i \in K_{e_t}}(\frac{1}{\mu}\max[0, 1-(m_{ti}- \overline{m}_{ti})(a_t - \overline{a}_t)]^2)
\end{aligned}
\end{gather}
where $\mu$ is a smoothing parameter which can be artificially adjusted. The square operation is designed to prevent non-differentiability.

\subsection{Predict Students' Responses}
To incorporate students' response labels into model training, we predict students' responses based on the difference between their knowledge mastery values and the difficulty levels of the target questions based on MIRT \cite{mirt}. The process is illustrated below:
\begin{equation}
\hat{y} = \frac{1}{1+e^{-\boldsymbol{r}_{q}{\boldsymbol{M}} + \theta_{q}}}
\end{equation}
where $\boldsymbol{r}_{q} \in \mathbb{R}^{K}$ can be seen as the relationship between question $q$ and $K$ knowledge concepts. $\boldsymbol{M}$ denotes the student's mastery values of $K$ concepts. $\theta_{q}$ denotes the difficulty of question $q$. However, considering that the question difficulty in Section 3.1 is obtained by statistics from the dataset, this method may introduce noise when the question occurs less frequently. Therefore, as previous studies did \cite{mf_dakt}, we introduce a fully-connected layer to map the embedding vector of the question to the difficulty value, making it immune to the noise of statistical features:
\begin{gather}
\widetilde{ \theta}_{q_{t+1}} = \sigma(\boldsymbol{W}_{6}^T \boldsymbol{q}_{t+1} + \boldsymbol{b}_{6}) 
\end{gather}
where $\widetilde{ \theta}_{q_{t+1}}$ denotes the estimated difficulty of question $q_{t+1}$ by our model. $\boldsymbol{W}_{6} \in \mathbb{R}^{d \times 1}$ and $\boldsymbol{b}_{6} \in \mathbb{R}^{1}$ are parameters to learn. 

In reality, when answering a target question, students do not need to have high mastery of concepts that are unrelated to the question. Hence, it is necessary to consider the relationship between the target question and concepts. Since manually annotated relationships may miss related concepts \cite{neuralcdm}, CMKT further mines the hidden relationship between the question and concept as follows:
\begin{gather}
r_{q_{t+1}, k_i} = \frac{\sigma({\boldsymbol{q}_{t+1}}^T (\boldsymbol{W}_{r}^T \boldsymbol{k}_{i}))+ \delta(k_i \in K_{q_{t+1}})}{\begin{matrix} \sum_{1 \leq j \leq N_k} [\sigma({\boldsymbol{q}_{t+1}}^T (\boldsymbol{W}_{r}^T \boldsymbol{k}_{j})) + \delta(k_i \in K_{q_{t+1}})] \end{matrix}}
\end{gather}
where $r_{q_{t+1}, k_i} \in \mathbb{R}^1$ is the relation score between $q_{t+1}$ and $k_i$. $\boldsymbol{W}_{r} \in \mathbb{R}^{d_k \times d_q}$ is the parameter to learn. Considering that concepts labeled by experts should have higher confidence, we add a function $\delta(k_i \in K_{q_{t+1}})$ which will be 1 if $q_{t+1}$ involves $k_i$ labeled by experts, otherwise 0, to give labeled concepts higher relation scores. Finally, we can obtain relation scores between $q_{t+1}$ and $N_k$ concepts: $\boldsymbol{r}_{q_{t+1}} =\left\{r_{q_{t+1}, k_1}, ..., r_{q_{t+1}, k_i}, ..., \right\}$.  After that, a binary cross-entropy loss is applied to directly optimize the assessed mastery of concepts:
\begin{gather}
\hat{y}_{t+1} = \frac{1}{1+e^{-\boldsymbol{r}_{q_{t+1}}^T{\boldsymbol{M}_t} + \widetilde{\theta}_{q_{t+1}}}} \\
\mathcal{L} = -\sum_{t=1}^T((y_t \log \hat{y}_{t} + (1-y_t)\log(1- \hat{y}_{t})))
\end{gather}

\subsection{Model Training}
In Section 4.4, we propose to use the question embedding vector to calculate the difficulty of the question. To expedite model convergence and ensure that the model learns the meaning of difficulty, we introduced an additional regularization term to constrain model learning. This regularization term, along with the prediction loss and the counterfactual monotonicity regularization term discussed earlier, constitutes the final objective function of the model:
\begin{gather}
\Phi_{ \theta} = \mathbb{E}[(\widetilde{ \theta}_{q}-{\theta}_{q})^2] \\ 
\mathcal{L}_{CMKT} := \mathcal{L} + \Phi_{CM} + \Phi_{\theta}
\end{gather}

\section{Experiments}

In this section, we conduct experiments with the aim of answering the following three research questions: \\
 \textbf{RQ1} How does CMKT perform compared to the existing KT or CD methods on assessing students' dynamic mastery of concepts? \\
 \textbf{RQ2} How do different components affect the performance of CMKT?
 \textbf{RQ3} How does counterfactual restriction improve KT baselines?

\subsection{Datasets} 
\textbf{\emph{ASSIST2009}}\footnote{https://sites.google.com/site/assistmentsdata/home\label{web}} is a popular public dataset collected from the ASSISTments platform  \cite{assist09}, which contains 269,009 samples from 2,983 students, a total of 109 knowledge concepts and 16,847 questions. A question involves four concepts at most. \textbf{\emph{ASSIST2012}}\textsuperscript{\ref {web}} also comes from the ASSISTments platform, but has 28,325 students and 2,710,820 practice records, involving 265 concepts and 53,079 questions, where each question is related to one concepts. \textbf{\emph{ASSIST2017}} \footnote{https://sites.google.com/view/assistmentsdatamining/dataset} comes from the 2017 ASSISTments data mining competition, with a total of 942,814 samples involving 1,709 students, 103 concepts, and 3,163 questions. Each question only involves one concept.  \textbf{\emph{EdNet}}\footnote{https://github.com/riiid/ednet\label{web2}} is collected from the Santa platform \cite{ednet}. We use a preprocessed smaller dataset from previous studies containing 222,141 records from 5,002 students with 187 concepts and 10,779 questions \cite{pebg}. A question can involve six concepts at most. \textbf{\emph{Algebra2005}}\footnote{https://pslcdatashop.web.cmu.edu/KDDCup/downloads.jsp\label{web3}} stems from the KDD Cup 2010 EDM Challenge, with 567 students, 172,994 questions and 111 concepts, and 606,359 records \cite{algebra05}, where each question involves one to seven concepts. Following previous CD works, we can calculate the mean standard deviation of scores that the student got for questions related to concepts to analyze whether students have large changes in their knowledge states as follows:
\begin{gather}
{STD}_{\# log>1}={mean}{[{mean}{[{std}_{ij}]}_{1 \le j \le N_k, \#log(i, j) > 1}]}_{1 \le i \le S}
\end{gather}
where $\#log(i, k)$ is the amount of practices of $s_i$ answered that involves concept $k_j$. $S$ denotes the total number of students.

\begin{table}[t]
	\caption{Statistics of five benchmark datasets.}
	\resizebox{\linewidth}{!}{
	\begin{tabular}{cccccc}
    		\hline Dataset & Students & Concepts & Questions & Records & ${STD}_{\#log>1}$ \\
		\hline ASSIST2009 & 2,983 & 109 & 16, 847 & 269,009 & 0.320 \\
		ASSIST2012 & 28,325 & 265 &53,079 & 2,710,820 & 0.304  \\
		ASSIST2017 & 1,709 & 103 & 3,163 & 942,814 & 0.445  \\
		EdNet & 5,002 & 187 & 10,779 & 222,141  & 0.350 \\
		Algebra2005 & 567 & 111 & 172,994 & 606,359  & 0.353 \\
		\hline
	\end{tabular}}
	\label{dataset}
\end{table} 

\subsection{Baselines}
We select the KT methods mentioned in their papers on how to obtain the knowledge mastery value as the baseline. \\
\textbf{$\bullet$ DKVMN} \cite{dkvmn}: It uses a key-value memory network to store and update students' knowledge states and can assess the mastery of concepts by setting the correlation weight to be [0, .., $w_i$=1 , ..0].\\
\textbf{$\bullet$ GKT} \cite{gkt}: It is the first to introduce graph into KT by initializing a concept graph optimized by predicting responses. GKT maintains a matrix which stores the vectors denoting mastering each concept. We can apply the sigmoid function to the sum of elements of each row vector of the matrix as the mastery value of each concept. \\
\textbf{$\bullet$ SKT} \cite{skt}: It extends GKT by exploiting the multiple relations in knowledge structure to model the influence propagation among concepts. We can use the method similar to GKT to obtain the mastery values of knowledge concepts assessed by SKT. \\
\textbf{$\bullet$ CKT} \cite{ckt}: It uses hierarchical convolutional layers to extract personalized learning rates and outputs a matrix that represents the student's mastery level of each concept. By applying the sigmoid function to the sum of the elements of each row vector of the output matrix, we can obtain the mastery value for each concept. \\ 
\textbf{$\bullet$ LPKT} \cite{lpkt}: It constrains a student's state evolution by measuring his learning gains between the current and previous practices. LPKT updates a knowledge mastery matrix for each student. We use the same method as CKT to obtain the mastery value of concepts.

To validate the performance in assessing concept mastery with CD methods, we also select several CD method as baselines: \\
\textbf{$\bullet$ NCDM} \cite{neuralcdm}: It incorporates the deep neural network to model complex interactions of students for getting accurate and interpretable diagnosis results with traditional Q-matrix. Each student can be mapped to an $K$-dimensional vector, where each element denotes the student's mastery of corresponding concept. \\
\textbf{$\bullet$ IRR} \cite{irr}: It incorporates monotonicity into the training process by creating sample pairs based on students' responses. We obtain the mastery value from the knowledge mastery vectors. \\
\textbf{$\bullet$ KSCD} \cite{kscd}: It aims at learning intrinsic relations among knowledge concepts from student response logs to interpretably infer the student's mastery over non-interactive concepts.

\subsection{Experimental Setup}
For all datasets, we performed standard 5-fold cross-validation on all models, that is, 80\% of the students' learning sequences were split into training and validation sets (their ratio was 8:2), and the remaining 20\% students are used as the test set. We set the sizes of $d_q$, $d_k$, $d_{q{\theta}}$, $d_{k{\theta}}$ , $d_a$ are all 64 and all input sequences to a fixed length of 200. We excluded sequence of students whose history records are longer than 10 because history exercise records that are too short are meaningless to predict. The difficulty levels of question and concepts, i.e. $C_q$ and $C_k$, are both set as 100. $\mu$ in the counterfactual restriction term is 0.25, which is a common setting. The optimizer is Adam \cite{adam} with a learning rate of 0.002.

\begin{table*}[t]
	\centering
	\caption{Results of models on five public datasets, where * indicates p-value \textless  0.05 in the significance test over the best baseline.}
	\begin{tabular}{c|c|c|c|c|c|c|c|c|c|c}
		\toprule
		Datasets & Metrics & DKVMN\cite{dkvmn} & GKT\cite{gkt} & SKT\cite{skt} & CKT\cite{ckt} & LPKT\cite{lpkt} & NCDM\cite{neuralcdm} & IRR\cite{irr}  & KSCD\cite{kscd} & CMKT \\
		\midrule 
		\multirow{3}*{ASSIST2009}  
		&ACC &0.7302 &0.7238 &0.7328  &0.7353&\underline{0.7401} &0.7149 &0.7183 &0.7237 &\textbf{0.7569*} \\ 
		&AUC &0.7481 &0.7436 &0.7516 &0.7634&\underline{0.7748} &0.7378&0.7402  &0.7451 &\textbf{0.7808*}  \\ 
		% &\textbf{\emph{GAUCM}} &0.5470 &0.5645 & 0.5617&0.5709&0.5898 &0.5906&0.6013  & 0.5993 &\textbf{0.6961*} \\
		\midrule
		\multirow{3}*{ASSIST2012}  
		&ACC &0.7359 &0.7373 &0.7472  &0.7395&\underline{0.7592} &0.7207&0.7325  &0.7384 &\textbf{0.7656*}  \\ 
		&AUC &0.7356 &0.7293 &0.7421 &0.7344&\underline{0.7799}  &0.7123&0.7267 &0.7316 &\textbf{0.7875*}  \\  
		&GAUCM &0.5432 &0.5736 &0.5653  &0.5751&0.5862 &0.5688&\underline{0.6291}  & 0.5843&\textbf{0.6948*} \\
		\midrule
		\multirow{3}*{ASSIST2017} 
		&ACC &0.6881 &0.6994 &0.7133  &0.7115&\underline{0.7184}  & 0.6542&0.6591&0.6679 &\textbf{0.7239*}  \\ 
		&AUC &0.7195 &0.7293 &0.7506  &0.7412&\underline{0.7759}  &0.6878 &0.6982&0.7012 &\textbf{0.7793*}  \\ 
		&GAUCM &0.5753 &0.5872 &0.5814 &0.5901&0.6094 &0.6023 &\underline{0.6382} & 0.6120 &\textbf{0.7126*} \\ 
		\midrule
		\multirow{3}*{EdNet} 
		&ACC &0.6806 &0.6812 &\underline{0.6923}  &0.6823&0.6912 &0.6554 &0.6685 & 0.6697 &\textbf{0.7005*}  \\ 
		&AUC &0.7214 &0.7240 &\underline{0.7343} &0.7269&0.7331 &0.6990 &0.7195 & 0.7132 &\textbf{0.7418*}  \\ 
		% &\textbf{\emph{GAUCM}} &0.5412 &0.5458 &0.5523 &0.5579&0.5634  &0.5718 &0.5691 &0.5843 &\textbf{0.6485*} \\ 
		\midrule
		\multirow{3}*{Algebra2005}&ACC &0.8015 &0.8092 &0.8114 &0.8113&\underline{0.8123} &0.8017 &0.8055 &0.8032 &\textbf{0.8179*}  \\ 
		&AUC &0.8028 &0.8125 &0.8161 &0.8168&\underline{0.8201} &0.7480 &0.7609  &0.7651 &\textbf{0.8263*}  \\ 
		% &\textbf{\emph{GAUCM}} &0.5372 &0.5419 &0.5528 &0.5601&0.5687 &0.5393 &0.5723  & 0.5471&\textbf{0.6471*} \\ 
		\bottomrule
	\end{tabular}
	\label{overall}
	\vspace{-0.3cm}
\end{table*} 

\subsection{Evaluation Metric}
Due to the lack of labels for students' mastery of concepts, we adopt the accuracy (ACC)  and area under the curve (AUC) as evaluation metrics. However, the black-box nature of deep networks and lack of explicit constraints on mastery values means that increased accuracy in predicting students' exercise answers does not necessarily translate to higher accuracy in assessing knowledge concept mastery. Therefore, we design a metric in this paper to directly and effectively quantify the accuracy of assessing concept mastery.

Monotonicity theory states that students who answer questions correctly should have a higher mastery level of the related concepts than students who answer incorrectly. Inspired by GAUC \cite{gauc}, this paper designs a metric called \textbf{Group AUC of Mastery (GAUCM)} which measures a KT method's ability to provide positive students (who answer the question correctly) with a higher level of concept mastery than negative students (who answer the question incorrectly) for each question. It's important to note that if a question involves multiple concepts, we cannot compare students' mastery of each concept based on their responses. This is because students who answered the questions incorrectly may not necessarily have a lower mastery value of each concept than the students who answered correctly. The wrong answer may simply indicate a poor mastery of a certain knowledge concept. Therefore, GAUCM will only be used in datasets where each question involves only one concept\footnote{The situation that a question involves multiple concepts is left for future research}. The calculation process for GAUCM is to compute the AUC of the mastery values for the concept involved in each question for all students who answered the question as follows:
\begin{gather}
\mbox{\emph{\textbf{GAUCM}}} = \frac{\begin{matrix} \sum_{q_j \in Q} N(q_j) \times AUC\Big[{\left \{ m_{ti, s} \right \}}_{0 \le t \le T, K_{q_j}={k_i}}^{s \in S_{q_j}}\Big] \end{matrix}}{\begin{matrix} \sum_{q_j \in Q} N(q_j) \end{matrix}}
\end{gather}
where $m_{ti, s}$ denotes student $s$'s mastery of $k_i$. $K_{q_j}=k_i$ denotes $k_i$ is labeled related to $q_j$ by experts. $S_{q_j}$ denotes the set of students who ever answered $q_j$. $Q$ denotes the question pool in the dataset. $N(q_j)$ is the number of $q_j$ answered and is utilized to average the AUC of different questions as GAUCM. A higher GAUCM denotes a stronger ability to give students more accurate mastery values.  

\subsection{Evaluation of Students' Mastery (RQ1)}
For all experiments, we deploy a t-test \cite{t-test} under the ACC metric, and a Mann-Whitney U test \cite{m-test} under AUC and GAUCM.  Table \ref{overall} presents all methods' performance, where the best results are bolded and the best baseline is underlined. Table \ref{overall} shows that our model, CMKT, outperforms the best baselines on the five datasets in terms of both ACC and AUC, demonstrating its reliability in predicting student responses to practices. Furthermore, since each question in the ASSIST2012 and ASSIST2017 datasets only relates to one concept, we can use the GAUCM metric to evaluate model performance in assessing the knowledge mastery. The results show that the CD methods perform worse than the KT methods in predicting student responses, likely due to the CD method's inability to capture the dynamics of learning. As shown in Table \ref{dataset}, the ASSIST2017 dataset has the largest $STD_{\# log > 1}$, indicating that students' knowledge state changes faster, leading to the worst performance of CD models. Despite the interpretable structure of the CD methods, their GAUCM is higher than that of the KT methods, but overall, it is not very high. This is because NCDM and KSCD also use deep neural networks for prediction and cannot satisfy monotonicity if they retrieve only the mastery of the target concept. IRR uses MIRT to predict students' responses, similar to our model, and outperforms the other two CD models. However, IRR is limited by the static knowledge state assumption and cannot dynamically trace students' knowledge states. Additionally, IRR still applies monotonicity theory at the question level, making it challenging to deal with situations involving multiple knowledge concepts in a question. Therefore, this highlights the need for exploring KT model improvements to assess students' mastery of concepts, rather than relying on the CD model for student concept mastery assessment. Moreover, compared to other KT baselines, LPKT enhances the ability to assess students' mastery of concepts by measuring learning gain between practices in a fine-grained manner, resulting in a more reasonable evolution process of knowledge states. This underscores the importance of modeling the evolution process of students' knowledge states. However, LPKT does not compare differences between learning gains generated by different answers. As a result, it falls short of satisfying the monotonicity of the evolution of concept mastery.

\begin{figure}[t]
  \centering
  \includegraphics[width=\linewidth]{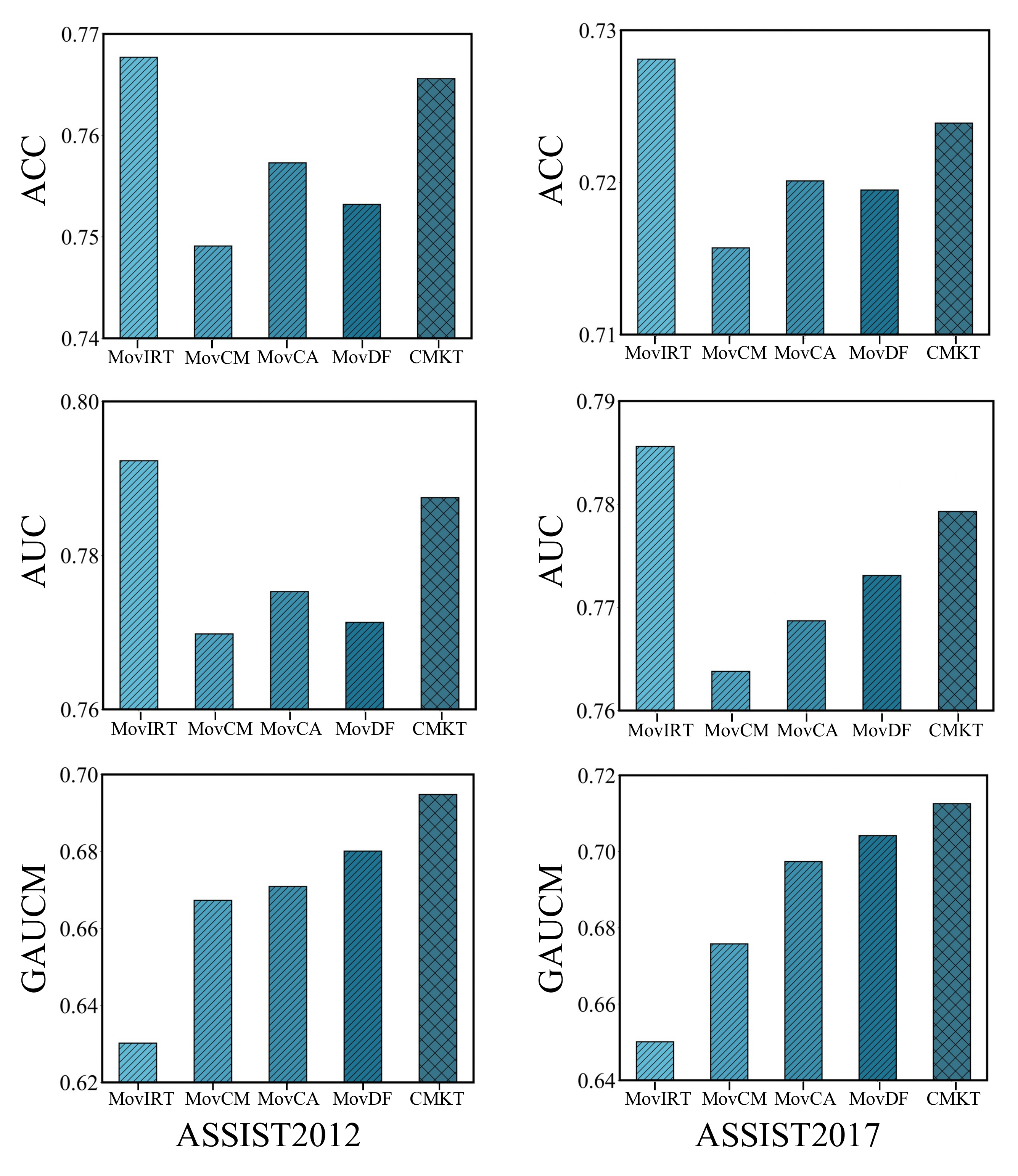}
  \caption{The experimental results of ablation study.}
  \label{ablation}
\end{figure}

\begin{table*}[t]
	\caption{The results of compatibility study, where * indicates p-value \textless  0.05 in the significance test over the original model.}
	\small
	\begin{tabular}{c|cc|ccc|ccc|cc|cc}
		\toprule
		\multirow{2}*{Model} & \multicolumn{2}{c|}{ASSIST2009}  & \multicolumn{3}{c|}{ASSIST2012} & \multicolumn{3}{c|}{ASSIST2017} & \multicolumn{2}{c|}{EdNet}  & \multicolumn{2}{c}{Algebra2005} \\
		& ACC &AUC & ACC & AUC & GAUCM & ACC & AUC & GAUCM &ACC & AUC &ACC & AUC \\
		\midrule 
		DKVMN &0.7302 &0.7481 &0.7359  &0.7356&0.5432 &0.6881 &0.7195 &0.5753 &0.6806 &0.7214 & 0.8015&0.8028 \\ 
		DKVMN + CM &0.7323* &0.7510* &0.7392* &0.7389*&0.5578* &0.6901* &0.7231* &0.5802* &0.6839* &0.7241* & 0.8056*&0.8099* \\ 
		\midrule
		GKT &0.7238 &0.7436 &0.7373  &0.7293 &0.5736 &0.6994 &0.7293 &0.5872 &0.6812 &0.7240 & 0.8092&0.8125 \\ 
		GKT + CM &0.7249* &0.7448* &0.7399*  &0.7357*&0.5792* &0.7013 *&0.7317* &0.5925* &0.6837* &0.7271* & 0.8124*&0.8151* \\ 
		\midrule
		SKT &0.7328 &0.7516 &0.7472  &0.7421&0.5653 &0.7133 &0.7506 &0.5814 &0.6923 &0.7343 & 0.8114&0.8161 \\ 
		SKT + CM &0.7335* &0.7529* &0.7491*  &0.7450*&0.5728* &0.7167* &0.7542* &0.5896* &0.6941* &0.7355*& 0.8142*&0.8189* \\ 
		\midrule
		CKT &0.7353 &0.7634 &0.7395  &0.7344&0.5751 &0.7115 &0.7412 &0.5901 &0.6823 &0.7269 & 0.8113&0.8168 \\ 
		CKT + CM &0.7392* &0.7691* &0.7469*  &0.7412*&0.5939* &0.7151* &0.7503* &0.6083* &0.6845* &0.7291* & 0.8154*&0.8182* \\ 
		\midrule
		LPKT &0.7401 &0.7748 &0.7592  &0.7799&0.5862 &0.7184 &0.7759 &0.6094 &0.6912 &0.7331 & 0.8123&0.8201 \\ 
		LPKT + CM &0.7421* &0.7762* &0.7623* &0.7841*&0.6343* &0.7201* &0.7773* &0.6416* &0.6928* &0.7351* & 0.8134*&0.8229* \\ 
		\bottomrule
	\end{tabular}
	\label{compa}
	\vspace{-0.3cm}
\end{table*} 

\begin{figure*}[t]
  \centering
  \includegraphics[width=0.9\linewidth]{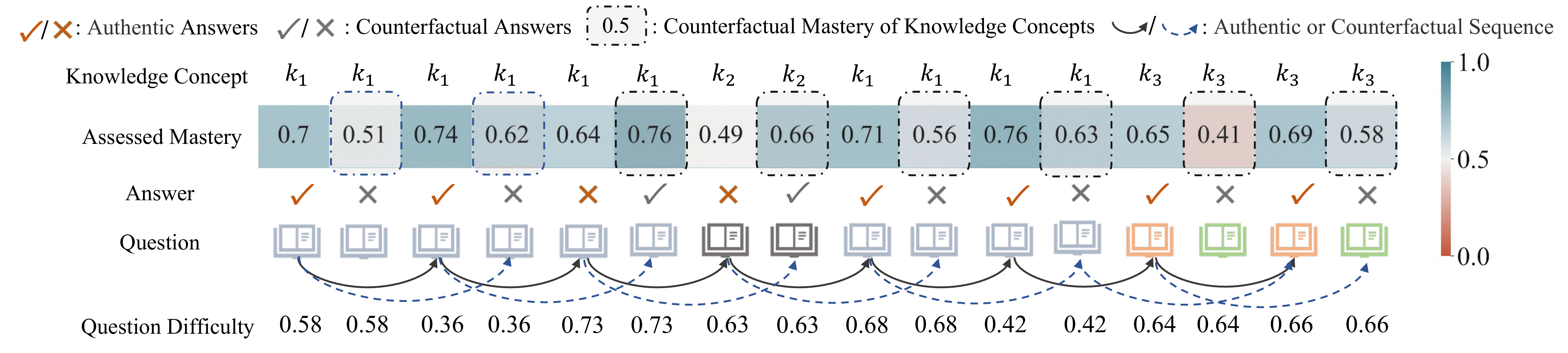}
  \caption{A case study of students' mastery of knowledge concepts on the ASSIST2012 dataset. $k_1$ \emph{: Addition and Subtraction Positive Decimals.} $k_2$\emph{: Ordering Positive Decimals.} $k_3$\emph{: Multiplication and Division Positive Decimals.}}
  \label{master}
  \vspace{-0.3cm}
 \end{figure*}

\subsection{Ablation Study (RQ2)}
We conduct the ablation study with for comparative settings:\\
\textbf{$\bullet$ MovCM} removes the counterfactual restriction function.   \\ 
\textbf{$\bullet$ MovCA} removes the virtual question constructed by concept difficulty (Eq. 7) and replace it with a randomly initialized vector. \\
\textbf{$\bullet$ MovDF} removes the regularization term of questions' difficulty. \\
\textbf{$\bullet$ MovIRT} removes MIRT and replace it with a deep network. 

Figure \ref{ablation} shows the average results of five runs, indicating that all variants reduce CMKT's performance in both datasets, with MovCM and MovIRT having the greatest impact on the GAUCM. The counterfactual constraint function (MovCM) can indicate the monotonicity of student learning progress, which is difficult to achieve with the prediction loss. By replacing MIRT with deep neural networks (MovIRT), the model's predictive performance improves, but the interpretability of concept mastery evaluation suffers. Thus, it's important to balance the predictive performance of the model with the reasonableness of the assess mastery value. Although removing MIRT improves the model's prediction performance, it also decreases the GAUCM significantly. As such, we believe it's worth sacrificing some predictability to obtain more reliable mastery results. MovCA causes more significant performance degradation of CMKT on ASSIST2012 than on ASSIST2017 due to the larger number of concepts in ASSIST2012. Eq. 7 connects the model's input (i.e., representation of $p_t$) with the virtual question representation to expedite model convergence. Additionally, removing the difficulty regularization term of questions harms the model's performance, but because ASSIST2017 has sufficient practice records per question, questions can still learn highly generalized representations, resulting in minor performance degradation.

\subsection{Compatibility Study (RQ3)}

Besides, we conduct compatibility study and present the results in Table \ref{compa}, where +CM refers to the application of the counterfactual monotonicity constraints during the baseline models training. Table \ref{compa} shows that CM effectively improves GAUCM, as well as ACC and AUC metrics in the baselines, and LPKT shows greater performance improvement than other baselines. We believe that the reason is that the motivation of counterfactual monotonicity is similar to modeling the learning gain between two practices, and the CM loss complements LPKT's design to achieve better performance. Additionally, we observe that all baselines still perform worse than CMKT after adding CM. We think this is because the prediction loss and structure of CMKT are specifically designed for monotonicity, enabling the CM to play a larger role. While the baselines use the deep network to predict responses, adding CM may lead to potential gradient conflicts. Therefore, it's essential to design a reasonable structure for assessing students' mastery of knowledge concepts, rather than merely introducing a specific training objective.

\subsection{Visualization}
To further comprehend students' mastery of concepts, we randomly selected an 8-question, 3-concept student record from the ASSIST2012 dataset, as shown in Figure \ref{master}. We can find that the student's mastery of $k_1$ is 0.7 when he correctly answer a question involving $k_1$ for the first time, which is higher than the mastery of counterfactual answers at 0.51. This phenomenon persists throughout the evolution process, verifying our model's effectiveness in learning counterfactual monotonicity. Additionally, Figure \ref{master} shows that a student's mastery of the same concept varies when practicing different question involving that concept. This variation may be due to differences in question difficulty. If a student correctly answers more difficult questions, his learning gain should be greater than that of less difficult questions, resulting in a more significant improvement in concept mastery. That's also why we should consider difficulty information in the knowledge extraction module.

\section{Conclusions}
In this paper, we propose a principled approach called Counterfactual Monotonic Knowledge Tracing (CMKT), which uses a type of counterfactual monotonicity to enhance the interpretability and semantic of assessed knowledge mastery values. Specifically, we first map students' historical practice sequences to their concept mastery values. Then, by constructing counterfactual sample pairs and conducting pair-wise training, we can directly constrain the knowledge update process of students in the model at the concept level to ensure semantic information of the assessed knowledge mastery value. Meanwhile, to take full advantage of student response label information, we predict their responses by measuring the difference between their mastery value and the difficulty level of the question. A prediction loss of students' responses is then used to facilitate model learning, together with the above counterfactual monotonic regularization term and a question difficulty regularization term. Finally, extensive experiments are conducted to verify the superiority of CMKT on five datasets based on a new metric called GAUCM and two commonly used ones.

\begin{acks}
This work is supported by 2022 Beijing Higher Education ``Undergraduate Teaching Reform and Innovation Project'' and 2022 Education and Teaching Reform Project of Beijing University of Posts and Telecommunications (2022JXYJ-F01).
\end{acks}

\end{document}